\documentclass{article}
\usepackage{spconf,amsmath,graphicx}
\usepackage{amssymb}
\usepackage[table]{xcolor}

\def\x{{\mathbf x}}

\title{Designing an Effective Metric Learning Pipeline for Speaker Diarization}
\name{Vivek Sivaraman Narayanaswamy$^*$, Jayaraman J. Thiagarajan$^{\dagger}$\thanks{This work was supported in part by the ASU SenSIP center, Arizona State University.  Portions of this work were  performed under the auspices of the U.S. Department of Energy by Lawrence Livermore National Laboratory under Contract DE- AC52-07NA27344. }, Huan Song$^{\ddagger}$, and Andreas Spanias$^*$}
\address{$^*$Arizona State University, $^\dagger$Lawrence Livermore National Labs, $^\ddagger$Bosch Research North America \\
Email:\{vnaray29@asu.edu, jjayaram@llnl.gov, huan.song@us.bosch.com, spanias@asu.edu\}}
\begin{document}
%
\maketitle
\begin{abstract}
State-of-the-art speaker diarization systems utilize knowledge from external data, in the form of a pre-trained distance metric, to effectively determine relative speaker identities to unseen data. However, much of recent focus has been on choosing the appropriate feature extractor, ranging from pre-trained $i-$vectors to representations learned via different sequence modeling architectures (e.g. 1D-CNNs, LSTMs, attention models), while adopting off-the-shelf metric learning solutions. In this paper, we argue that, regardless of the feature extractor, it is crucial to carefully design a metric learning pipeline, namely the loss function, the sampling strategy and the discrimnative margin parameter, for building robust diarization systems. Furthermore, we propose to adopt a fine-grained validation process to obtain a comprehensive evaluation of the generalization power of metric learning pipelines. To this end, we measure diarization performance across different language speakers, and variations in the number of speakers in a recording. Using empirical studies, we provide interesting insights into the effectiveness of different design choices and make recommendations.
\end{abstract}
\section{Introduction}
\label{sec:intro}
Speaker diarization refers to the problem of attributing relative speaker identities without any prior information about speakers or the nature of speech~\cite{tranter2006overview}. This is often used as the first step before invoking downstream inference tasks such as speaker recognition. Diarization systems can be severely challenged by variabilities in acoustic conditions and the need to adapt to speakers with different characteristics. Posed as an unsupervised learning problem, its success relies heavily on the choice of an appropriate distance metric for performing clustering. While classical approaches~\cite{prazak2011speaker,sell2014speaker,anguera2012speaker} resorted to careful feature design coupled with a predefined metric, for example cosine similarity between $i-$vectors, more recent solutions have emphasized the importance of integrating a metric learning pipeline into diarization systems~\cite{garcia2017speaker,le2017triplet,song2018triplet}. This naturally allows knowledge inferred from an external data source to be utilized while performing diarization on an unseen target data. Powered by recent advances in deep neural networks, there is a surge in interest to construct generalizable latent spaces, that will make the learned metric highly effective for even unseen speakers~\cite{cyrta2017speaker}.
 
In general, metric learning aims to utilize latent features in data to effectively compare observations~\cite{hoffer2015deep}. This amounts to inferring key factors in data, while encoding higher order interactions, to ensure that examples from the same speaker are within smaller distances, compared to examples from a different speaker~\cite{hoffer2015deep}. While a variety of formulations exist for supervised metric learning~\cite{song2018triplet,wang2017speaker}, recent approaches have relied on deep networks to construct embeddings that satisfy the supervisory constraints. Popular examples include the \textit{siamese}~\cite{koch2015siamese}, \textit{triplet}~\cite{schroff2015facenet,hoffer2015deep}, and \textit{quadruplet}~\cite{chen2017beyond} networks. By coupling sequence modeling techniques with these deep metric learning formalisms, recent works such as~\cite{garcia2017speaker,song2018triplet} produce state-of-the-art diarization performance, while entirely dispensing the need for explicit feature design. 

Though most existing works have focused on choosing the right sequence modeling architecture, it is critical to understand the impact of different components in the metric learning pipeline, from the context of generalization performance. In this paper, we consider three critical components in deep metric learning algorithms, and perform empirical studies to understand their impact on diarization performance: (i) loss function, (ii) strategy for sampling negative examples, and (iii) margin parameter selection. By performing a fine-grained evaluation of generalization to different language speakers and variations in the number of speakers, our study provides interesting insights into choosing the right metric learning architecture for reliable performance.

\section{Diarization System Overview}
\label{sec:arch}
\begin{figure*}[t]
	\centering
	\includegraphics[width=0.75\textwidth]{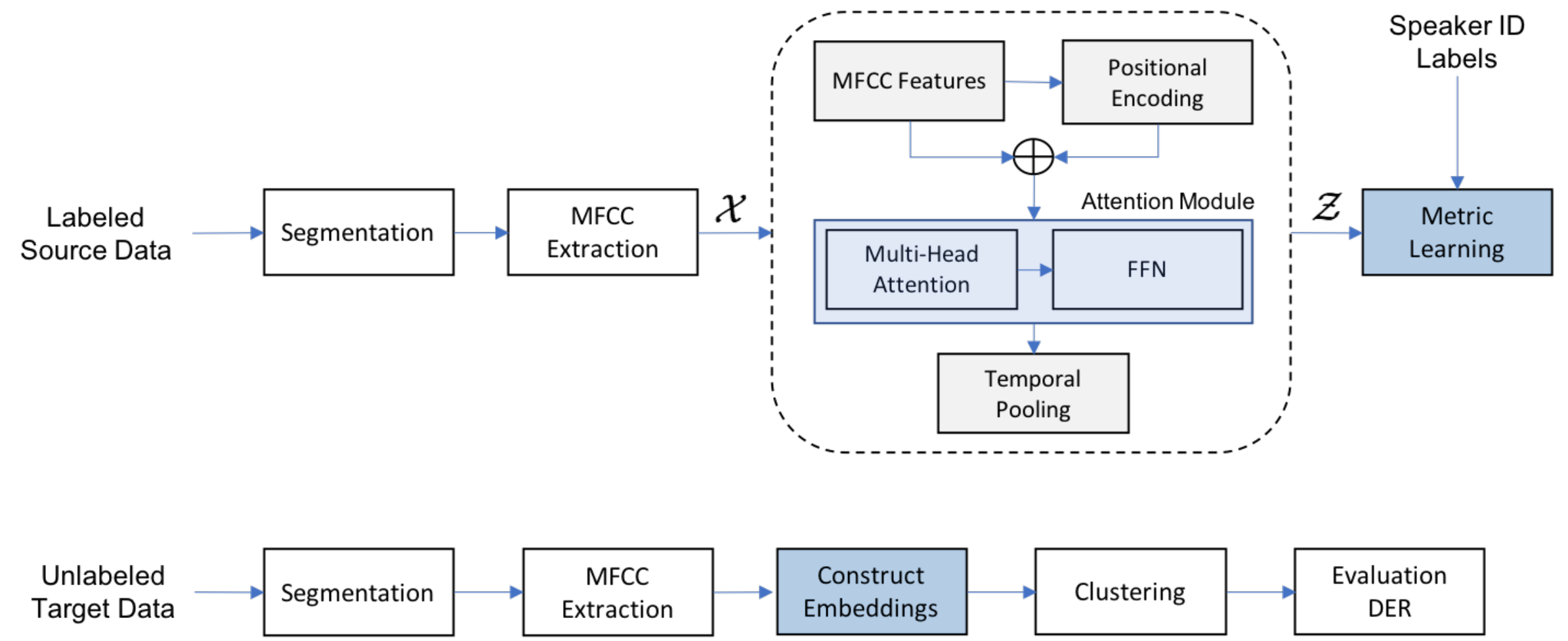}
	
	\caption{An overview of the diarization system adopted in this work. Following the state-of-the-art approach in~\cite{song2018triplet}, we use raw MFCC features along with deep metric learning to infer embeddings for diarization. The focus of this work is to effectively design different components of metric learning, such that improved generalization is achieved.}
	\vspace{-0.1in}
	\label{fig:approach}
\end{figure*}

An overview of the diarization system used in our work is illustrated in Figure \ref{fig:approach}. Though, several existing solutions build upon pre-designed features, such as $i-$vectors, our setup follows the approach in~\cite{song2018triplet} and operates directly on the mel frequency cepstral coefficients (MFCCs) to extract speaker embeddings. In the first stage, an out-of-domain labeled source dataset is utilized to perform metric learning, wherein the speaker ID is used to define positive and negative triplets (or quadruplets). For learning latent features from the sequence data, we adopt the state-of-the-art attention models~\cite{vaswani2017attention}. Diarization is then performed on a different target dataset by first extracting embeddings with the pre-trained model and then organizing segments using a clustering algorithm. 

\noindent \textbf{Preprocessing:} In our setup, the speech recordings considered are temporally segmented into non-overlapping segments of equal duration (fixed at $2$ seconds). The MFCC features are then extracted using $25$ms Hamming windows with $15$ms overlap. Consequently, each data sample corresponds to a temporal sequence $\mathbf{x}_i$ $\in$ $\mathbb{R} ^{T \times d}$ where $T$ is the number of MFCC frames and $d$ is the number of MFCC dimensions. Similar to ~\cite{song2018triplet}, $d$ was chosen to be 60.

\noindent \textbf{Architecture:} In this work, attention models are used to learn speaker embeddings from MFCC features, using a metric learning objective. Attention mechanism is a widely-adopted strategy in sequence modeling, wherein a parameterized function is used to determine relevant parts of the input to focus on, in order to make decisions~\cite{xu2015show,hermann2015teaching}. We use a popular implementation of attention models, \textit{Transformer}~\cite{vaswani2017attention}, which employs the scalar dot-product attention mechanism. 

This architecture uses a \textit{self-attention} mechanism to capture dependencies within the same input and employs multiple attention heads to enhance the modeling power. One useful interpretation of self-attention is that it implicitly induces a graph structure for a given sequence, where the nodes are time-steps and the edges indicate temporal dependencies. Furthermore, instead of a single attention graph, we can actually consider multiple graphs corresponding to the different attention heads, each of which can be interpreted to encode different types of edges and hence can provide complementary information about different types of dependencies. This concept is referred to as using \textit{multiple attention heads}. Song \textit{et al.}~\cite{song2018triplet} utilized a variant of this architecture for speaker diarization and our system follows their implementation. As illustrated in Figure \ref{fig:approach}, the model consists of a multi-head, self-attention mechanism with a feed-forward network (FFN) stacked together $L$ times to learn the deep representations. Besides, positional encoding is included to exploit the ordering information from a sequence. 


\noindent \textbf{Clustering:} After obtaining the embeddings $\mathcal{Z}$ from the pre-trained model from the out-of-domain data, we perform x-means~\cite{pelleg2000x} to estimate the number of speakers, and then use $k$-means clustering with the estimation. Note, we force x-means to produce at least $2$ clusters. 

\noindent \textbf{Evaluation Metric:} Following standard practice, we use diarization error rate (DER) as the evaluation metric and utilize the \texttt{pyannote.metric}~\cite{bredin2017pyannote} package. 

\section{Design Methodology}
\label{sec:design}
In this section, we describe design choices that we considered pertinent to loss function, the sampling strategy and selection of the margin. The design choices on these components result in a total of $11$ realizations of the metric learning pipeline.

\subsection{Choice of loss function}
\label{sec:loss}

We consider two state-of-the-art loss functions to build speaker embeddings from MFCC features with attention models: triplet loss, and quadruplet loss. Denoting the attention model as $\mathcal{A}(.)$, and the Euclidean distance between a pair of embeddings as $D_{ij}=\lVert \mathcal{A}(\x_i) - \mathcal{A}(\x_j) \rVert_2$, we describe the definitions of the loss functions in detail.

\noindent \textit{(i) Triplet loss (Trip)~\cite{hoffer2015deep}:} In a triplet network, every input to the attention model is a group of $3$ samples $\x_a, \x_p, \x_n$, where $\x_a$ denotes an anchor, $\x_p$ denotes a positive sample from the same class as $\x_a$, and $\x_n$ a negative sample from a different class. Every sample in the set is processed independently by the attention model, and we compute the triplet loss as:
\begin{equation}
\label{eq:triplet}
l_\text{trip}(\x_a, \x_p, \x_n)=\max(0, D_{ap}^2-D_{an}^2+\alpha),
\end{equation}where the margin parameter $\alpha$ characterizes the separation between $D_{an}^2$ and $D_{ap}^2$, such that $D_{an}^2\geq D_{ap}^2+\alpha$. Unlike the contrastive loss~\cite{koch2015siamese}, the triplet loss does not impose a global margin of separation, and allows a certain amount of distortion in the embedding space. 

\noindent \textit{(ii) Quadruplet loss (Quad)~\cite{chen2017beyond}:} A well-known criterion for achieving high generalization ability to unseen classes is to reduce the intra-class variability while enlarging the inter-class variability. The recent study on quadruplet network shows that, by adding such a modeling term into the triplet loss, one can decrease the generalization error~\cite{chen2017beyond}. More specifically, quadruplet loss includes an additional sample $\x_q$ to the input set of $\{\x_a, \x_p, \x_n\}$, where $\x_q$ is from a class different than both $\x_a$ and $\x_n$. As a result, the modeling of the intra- and inter-class variations can be achieved by targeting that $D_{qn}^2\geq D_{ap}^2+\alpha_2$, in addition to the triplet loss:
\begin{equation}
\begin{aligned}
	\label{eq:quadruplet}
	l_\text{quad}(\x_a, \x_p, \x_{n}, \x_{q})&=\max(0, D_{ap}^2-D_{an}^2+\alpha_1) \\
											 &+ \max(0, D_{ap}^2-D_{qn}^2+\alpha_2) 
\end{aligned}
\end{equation}
where $\alpha_1$ has the similar effect as in Equation \ref{eq:triplet} while $\alpha_2$ balances the two criteria in the training process.

\begin{figure*}[t]
	\centering
	\includegraphics[width=0.9\textwidth]{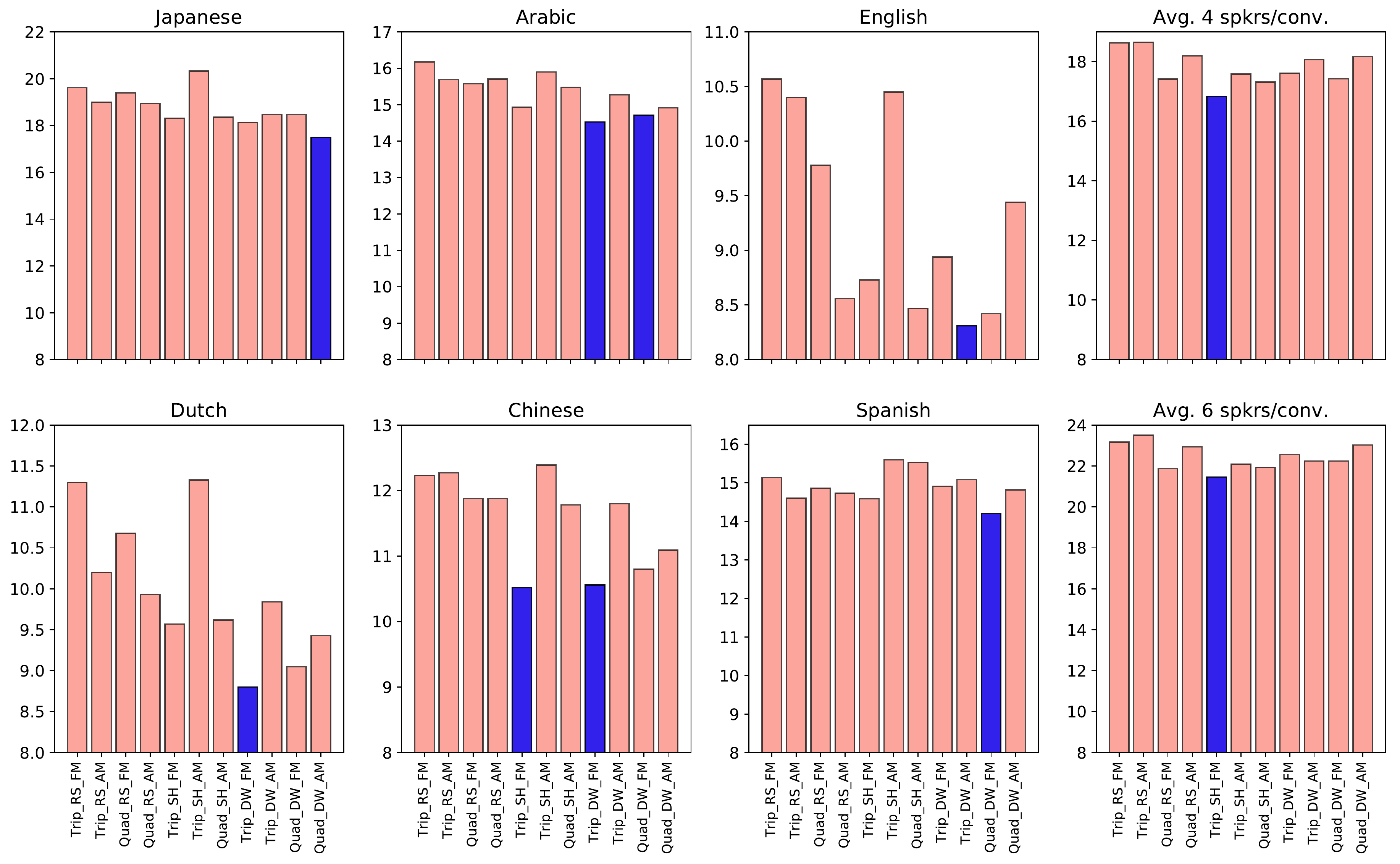}
	\label{fig:der_lang}
	\caption{Fine-grained evaluation of diarization performance. We measure the generalization power of a learned metric by performing diarization with speakers from a specific language, and also with increase in the average number of speakers.}
	\vspace{-0.1in}
	\label{fig:eval}
\end{figure*}

\subsection{Choice of sampling strategy}
\label{sec:sampling}
The sampling process is critical in training metric learning architectures, and recent studies have demonstrated that a good sampling strategy can be equally important as the loss formulation in achieving state-of-the-art performance~\cite{manmatha2017sampling}. We describe different sampling strategies under the setting of triplet loss, but the design choice is similar under other loss functions. Given an anchor-positive pair, the na\"ive way to obtain a negative sample is sample at random. However, this can be sub-optimal as a large number of random examples selected can be \textit{easy} negatives, which do not contribute to the loss at all. In this paper, we consider the following strategies:
\noindent \textit{(i) Semi-Hard Mining (SH)~\cite{schroff2015facenet}:} In constrast to random sampling, one can select only a \textit{hard} negative which satisfies $D_{an}^2\geq D_{ap}^2+\alpha$. However, as shown in~\cite{schroff2015facenet}， this can typically lead to a collapsed model. In order to construct more useful triplets (or quadruplets), it is prudent to select only those sets of embeddings that satisfy $D_{ap}^2\leq D_{an}^2\leq D_{ap}^2+\alpha$, referred as \textit{semi-hard} negatives. 

\noindent \textit{(ii) Distance-Weighted Sampling (DW)~\cite{manmatha2017sampling}:} Although semi-hard negative mining is effective in practice, it is still a heuristic approach and may not be optimal to cover the high-dimensional sampling space. Consequently, Wu \textit{et al.} analyzed existing sampling strategies, and hypothesized that a better approach can be to reduce the sampling bias, and be more exposed to classes which may lie at the edge of a latent space~\cite{manmatha2017sampling}. Specifically, we construct a discrete probability measure for each example based on the inverse distances to the anchor, and draw samples with the assigned probabilities. 

\begin{table}[t]
	\centering
	\caption{Overall performance of different configurations of metric learning. Our recommendations are showed in green.}
	\renewcommand{\arraystretch}{1.2}
	\begin{tabular}{|c|c|c|c|}
		
		\hline
		\cellcolor{gray!25}\textbf{Sampling} & \cellcolor{gray!25}\textbf{Loss} & \cellcolor{gray!25}\textbf{Margin} & \cellcolor{gray!25}\textbf{DER \%} \\ \hline \hline
		Random            & Triplet       & Fixed           & 14.11           \\ \hline
		Random            & Triplet       & Adaptive        & 13.57           \\ \hline
		Random            & Quadruplet    & Fixed           & 13.54           \\ \hline
		Random            & Quadruplet    & Adaptive        & 13.08           \\ \hline
		\cellcolor{green!30}Semi-hard         & \cellcolor{green!30}Triplet       & \cellcolor{green!30}Fixed           & \cellcolor{green!30}{12.77}  \\ \hline
		Semi-hard         & Triplet       & Adaptive        & 14.25           \\ \hline
		Semi-hard         & Quadruplet    & Adaptive        & 13.18           \\ \hline
		\cellcolor{green!30}DWS               & \cellcolor{green!30}Triplet       & \cellcolor{green!30}Fixed           & \cellcolor{green!30}{12.44}  \\ \hline
		\cellcolor{green!30}DWS               & \cellcolor{green!30}Triplet       & \cellcolor{green!30}Adaptive        & \cellcolor{green!30}12.98          \\ \hline
		\cellcolor{green!30}DWS               & \cellcolor{green!30}Quadruplet    & \cellcolor{green!30}Fixed           & \cellcolor{green!30}12.47           \\ \hline
		\cellcolor{green!30}DWS               & \cellcolor{green!30}Quadruplet    &\cellcolor{green!30} Adaptive        & \cellcolor{green!30}12.76           \\ \hline
	\end{tabular}
	\label{tab:results}
	\vspace{-0.1in}
\end{table}

\subsection{Choice of margin parameter}
\label{sec:margin}
Finally, we study the impact of how the margin parameter is chosen. In the \textit{fixed margin (FM)} case, pre-defined values were used throughout the training, $\alpha = 0.8$ ($\alpha_1 = 0.8$ and $\alpha_2 = 0.4$ for quadruplet loss). For the \textit{adaptive margin (AM)} case, as suggested in~\cite{chen2017beyond}, we compute margin as the difference between the mean of the anchor-negative distance distribution, $\mu_{an}$, and the mean of the anchor-positive distance distribution, $\mu_{ap}$, within every mini-batch processed by the network. During the initial phase of training, the margin is assigned to be a fixed value and as the training progresses, the margin increases and consequently only allows those samples producing a non-zero loss to be evaluated in the gradients computation. Our adaptive margin was calculated as follows:
\begin{equation}
\label{eq:adaptivemargin}
\alpha(\x_a, \x_p, \x_n)=\max(0.8, \mu_{an}-\mu_{ap}).
\end{equation}

\section{Empirical Evaluation and Inferences}
\label{sec:result}
All variants of the metric learning pipeline were trained on the TEDLIUM corpus which consists of $1495$ audio recordings. After ignoring speakers with less than $45$ transcribed segments, we have a set of $1211$ speakers with an average recording length of $10.2$ minutes. All recordings were down-sampled to $8$kHz to match the target CALLHOME corpus. The CALLHOME corpus consists of 780 transcribed, conversation speech recordings from six different languages namely Arabic, Chinese, English, German, Spanish and Japanese, containing 2 to 7 speakers.  We evaluate diarization performance using DER to understand impact of the different design choices on the overall diarization performance. 

Although DER collectively considers false alarms, missed detections and confusion errors, most existing systems evaluated on CALLHOME \cite{garcia2017speaker} account for only the confusion rate and ignore overlapping segments. Following this convention, we use the oracle speech activity regions and use only the non-overlapping sections. Additionally, there is a collar tolerance of $250$ms at both beginning and end of each segment.  Table \ref{tab:results} shows the overall diarization performance obtained using different realizations of the metric learning pipeline. By studying the impact of different design choices, we make the following observations: Both the triplet and quadruplet losses are highly effective in constructing generalizable latent spaces, however their performances are suboptimal when random sampling was used. On the other hand, distance weighted negative sampling boosts the performance significantly in all cases. However, semi-hard negative mining was useful only with the triplet loss. Finally, in contrast to state-of-the-art results in vision applications, using an adaptive margin did not seem to provide any improvements to the performance, except in the case of random sampling. Overall, we find that triplet loss with distance weighted sampling produced the lowest DER ($12.44\%$) on the CALLHOME dataset, which is the best reported performance in the literature so far.

Figure \ref{fig:eval} shows the language specific diarization performance of the architectures. It can be clearly observed that the DERs for English conversations are relatively lower than other languages. We attribute this to the fact the metric was trained using recordings in English. Another possible reason for higher DER in other languages can be related to the fact that speaker identity is the only semantic information used for training the network and as a result the model is not able to generalize the learned embeddings independent of language.

We also studied the effect of number of speakers in a conversation on the expected diarization performance - for this experiment, we randomly concatenated speech segments from multiple conversations, to produce two variants of the CALLHOME dataset where the average number of speakers per conversation was increased to $4$ and $6$ respectively. It can be observed from Figure \ref{fig:eval}, that the DER increases substantially for all the architectures considered, which clearly evidences the gaps in the current art of generalizing a distance metric to unseen scenarios.

In summary, we find that the choice of metric learning pipeline has a crucial role in diarization performance with unseen datasets, and we identify configurations that produce state-of-the-art results. However, we notice significant performance variability across speakers from different languages, and that the performance of diarization systems quickly degrades with increase in number of speakers per conversation.



%
%

\bibliographystyle{IEEEbib}
\bibliography{refs}

\end{document}